\documentclass[sn-basic]{sn-jnl}

\jyear{2021}%

\theoremstyle{thmstyleone}%
\theoremstyle{thmstyletwo}%

\theoremstyle{thmstylethree}%

\usepackage{graphicx}
\usepackage{amsmath,amsfonts,amssymb,commath,adjustbox}
\newcommand{\bftab}{\fontseries{b}\selectfont}

\usepackage[normalem]{ulem}
\useunder{\uline}{\ul}{}
\usepackage{subcaption}

\usepackage[normalem]{ulem}
\newcommand{\red}[1]{{#1}}

\begin{document}

\title{Language with Vision: a Study on Grounded Word and Sentence Embeddings}
\author{Hassan Shahmohammadi, Maria Heitmeier, Elnaz Shafaei-Bajestan, Hendrik P. A. Lensch, and R. Harald Baayen\\\\ University of Tübingen
\texttt{hassan.shahmohammadi@uni-tuebingen.de}}







\abstract{Grounding language in vision is an active field of research seeking to construct cognitively plausible word and sentence representations by incorporating perceptual knowledge from vision into text-based representations. Despite many attempts at language grounding, achieving an optimal equilibrium between textual representations of the language and our embodied experiences remains an open field. Some common concerns are the following. Is visual grounding advantageous for abstract words, or is its effectiveness restricted to concrete words? What is the optimal way of bridging the gap between text and vision? To what extent is perceptual knowledge from images advantageous
for acquiring high-quality embeddings? Leveraging the current advances in machine learning and natural language processing, the present study addresses these questions by proposing a simple yet very effective {computational grounding model for pre-trained word embeddings.} Our model effectively balances the interplay between language and vision by aligning textual embeddings with visual information while simultaneously preserving the distributional statistics that characterize word usage in text corpora. By applying a learned alignment, we are able to indirectly ground unseen words including abstract words. A series of evaluations on a range of behavioural datasets shows that visual grounding is beneficial not only for concrete words but also for abstract words, lending support to the indirect theory of abstract concepts. Moreover, our approach offers advantages for contextualized embeddings, such as those generated by BERT \citep{devlin2018bert}, but only when trained on corpora of modest, cognitively plausible sizes. Code and grounded embeddings for English are available at\footnote{Accepted in Behavior Research Methods Journal: \url{https://github.com/Hazel1994/Visually_Grounded_Word_Embeddings_2}}.}




\maketitle

\section{Introduction}
Where do symbolic representations of language get their meaning from? It has been argued both from a theoretical and an empirical perspective that knowledge is grounded in perceptual experience \citep{Barsalou2008GroundedCognition, Lakoff1987WomenThings, Langacker1999ALinguistics, Zwaan2005EmbodiedComprehension}. Evidence for this embodied view of knowledge comes from a range of scientific domains such as neuroimaging \citep[e.g.][]{Simmons2005PicturesReward, Martin2007TheBrain} and behavioural studies \citep[e.g.][]{Goldstone1995EffectsPerception, Solomon2001RepresentingLocally, Solomon2004PerceptualVerification}, showing that knowledge is grounded in sensory, but also interoceptive perception and motor action \citep[overview in][]{Barsalou2008GroundedCognition}. However, this view is not uncontested. For example, \citet{Louwerse2011AWords} argue that linguistic information suffices for more shallow processing of meaning and that perceptual, embodied information is only accessed when deeper knowledge of a word is required. 

This debate has been stimulated further by the success of meaning representations which are based on linguistic information alone. They build on the notion of \citet{Harris1954Distributional} that similar words occur in similar contexts and represent each word as numerical vectors, with similarities between these vectors reflecting similarities in words' meanings. By now, many different methods have been devised to generate such vectors \red{(called ``word embeddings'' in Natural Language Processing (NLP) and throughout the remainder of this paper)}, beginning with Hyperspace Analogue of Language \citep[HAL;][]{Lund1996ProducingCo-occurrence} and Latent Semantic Analysis \citep[LSA;][]{Landauer1997AKnowledge.}, and later, mainly in the \red{fields} of NLP and machine learning, Word2Vec \citep{Mikolov2013EfficientSpace}, Fasttext \citep{Bojanowski2017EnrichingInformation} or GloVe \citep{Pennington2014Glove:Representation}. Today, word embeddings are employed successfully in many different areas and tasks within NLP, such as POS-tagging, named-entity recognition, and sentiment analysis \citep{wang2019evaluating}.

As an easily obtained representation of semantics, word embeddings are also used in many areas of cognitive science, such as AI research, psychology or psycholinguistics, with encouraging results \citep[see][]{Gunther2019}. From a cognitive perspective, word embeddings have been evaluated in two ways. A relatively direct method is to compare them to metrics obtained from brain imaging such as fMRI or EEG. \citet{Bulat2017SpeakingBrain, Hollenstein2019CogniVal:Evaluation} showed that a variety of word embeddings (e.g. GloVe, Word2Vec, Fasttext) correlate relatively well with such metrics. A second, more indirect, approach uses behavioural data such as reaction times or ratings as evaluation criteria. \citet{mandera2017explaining} showed that word embeddings can be used to predict semantic priming as well as word associations, similarity/relatedness ratings and even perform well in a multiple-choice task. Further evidence in favour of the cognitive plausibility of word embeddings has been provided by \citet{Westbury2014YouJudgments, Westbury2019WrigglyFunny} who predicted familiarity and humour ratings respectively, \red{\citet{marelli2018database} who demonstrated that the semantic relatedness of words' orthographic neighbours is predictive for visual lexical decision and naming latencies}, \citet{Abdou2021CanColor} who showed that even color relations are accurately represented by purely textual embeddings, as well as \citet{louwerse2009greographical, avery2021reconstructing,gatti2022spatial} who demonstrated that geographical locations of cities are reflected in purely textual embeddings.  \red{Recently, embeddings have also found their way into psycholinguistic models. For example, the Discriminative Lexicon Model \citep{baayen2019discriminative, heitmeier2021modeling, heitmeier2023trial}, a model of the mental lexicon, uses word embeddings to represent words' meanings. Other models also use distributional information to represent semantics, either randomly generated ones \citep[e.g.][]{gaskell1997integrating, magnuson2020earshot} or based on human ratings \citep[e.g.][]{mirkovic2008acquisition}, further highlighting the need for a large set of psychologically valid word embeddings.} However, the cognitive plausibility of mechanisms generating word embeddings such as \textit{Word2Vec} has not gone unchallenged \citep{mannering2021catastrophic}.

While the success of textual embeddings has nevertheless led some researchers to believe that meaning can be fully, or at least to a large extent, be derived from language alone \citep{Landauer1999LatentAlone}, the wide range of empirical evidence in favour of a grounded view of knowledge representation and cognition has sparked the search for representations that are informed not only by text, but also by vision and other modalities \red{\citep[see also][]{andrews2014reconciling}}.

Therefore, a number of previous studies have tried to improve word embeddings by using available data similar to text corpora.  Some studies have tried to extract meaning representations exclusively from visual information (usually images). The resulting \red{visual word embeddings} have been found to be very good models of human perceptual behaviour \citep[e.g.][]{Zhang2018TheMetric}, but success at predicting other behavioural data was more mixed, with some reporting positive \citep{Luddecke2019DistributionalText, Bulat2017SpeakingBrain} and others negative results \red{compared to textual embeddings \citep[e.g.][]{Peterson2017AdaptingReport, DeDeyne2021VisualMind, rotaru2020constructing, utsumi2022test}}. The more promising approach has been to \red{ground textual embeddings in vision, i.e. to include visual information with textual embeddings. The resulting embeddings are usually referred to as multimodal embeddings}. This \red{approach} is especially promising because textual and visual representations seem to carry different kinds of information 
\red{\citep{Petilli2021Data-drivenProcessing, andrews2014reconciling}}. Multimodal embeddings have been successful in a range of areas. They have been shown to correlate better than purely textual embeddings with human similarity/relatedness \red{judgments} and concept categorization. \citet{Bulat2017SpeakingBrain, anderson2015reading} found that they are better at predicting brain activity than purely textual embeddings. Moreover, they are useful in modelling the learning of novel words' meanings in both children and adults \citep{Lazaridou2016MultimodalInput, Lazaridou2017MultimodalText}. \red{Finally, they have been shown to improve performance in a number of classification tasks in NLP \citep{bordes-etal-2019-incorporating}.}

Several approaches to obtaining multimodal embeddings are available. \red{We restrict our discussion here to approaches combining textual and visual information, but a body of work has also explored the integration of emotional \citep[e.g.][]{rotaru2020constructing}, sensory \citep[e.g.][]{Johns2012PerceptualSimilarity}, auditory \citep{kiela2015multi} and olfactory \citep{kiela2015grounding} information. Early approaches gleaned visual information from human ratings, e.g. by utilising data collected in the ESPGame dataset \citep{von2006games}, or used ``Bag-of-Visual-Word'' approaches where images are chunked into small pieces to form a kind of visual vocabulary \citep[e.g. in][]{anderson2015reading}. More recently, feature vectors have been extracted directly from computer vision models \citep[see][for a review]{baroni2016grounding}.}

\red{Subsequently, the visual information needs to be combined with textual information. \citet{baroni2016grounding} differentiates between two approaches: \textit{cross-modal mapping} and \textit{multimodal fusion}. Cross-modal mapping approaches to grounding textual in visual information aim to map between one and the other, in an attempt to account for how vision could be translated into language or vice versa \citep{baroni2016grounding}. An early model inferring perceptual embeddings by linking words using distributional semantics is \citet{Johns2012PerceptualSimilarity}. They used feature norms from \citet{McRae2005SemanticThings} to model perceptual representations. For words for which no feature norms were available, they inferred these by first computing the similarity of the target word with all words for which feature norms were available using distributional semantics, and then computed a weighted average of their feature norms. After having inferred feature norms for all words, they repeated the process in a second step, this time taking into account all words, rather than only those for which feature norms were available originally. A more recent proposal for connecting textual and visual embeddings by means of a simple linear mapping can be found in  \citet{Gunther2020Images}.}

\red{On the other hand, multimodal fusion \citep{baroni2016grounding} aims to combine textual and visual information into a single representation. The simplest example for multimodal fusion is concatenation, as is often used when multimodal embeddings are explored in cognitive science and psychology \citep[e.g.][]{utsumi2022test, rotaru2020constructing}. However, there are also more sophisticated approaches from the realm of NLP:}  Some approaches apply feature-level fusion, combining image features with textual word embeddings (after obtaining both separately) with methods such as \red{Singular Value Decomposition (SVD) or Gated Recurrent Units (GRU)}  \citep{cho2014properties, bruni2014multimodal,kiela-bottou-2014-learning, kiros-etal-2018-illustrative}.  Others learn multimodal word representations in a joint feature space defined by a specific criterion (known as loss function) between modalities, for example by using auto-encoders \citep{silberer-lapata-2014-learning,hasegawa-etal-2017-incorporating} or  \red{Long-Short Term Memory (LSTM)} \citep{hochreiter1997long} networks \citep{kiela-etal-2018-learning,chrupala-etal-2015-learning}. Recently, new approaches based on modality alignment have emerged. Here, vision and language are treated separately (as opposed to having both in a shared space) but the textual embeddings are aligned with image features \citep{shahmohammadi2021learning, bordes-etal-2019-incorporating}. 

\begin{figure}[hb]
  \centering
  \includegraphics[width=\textwidth]{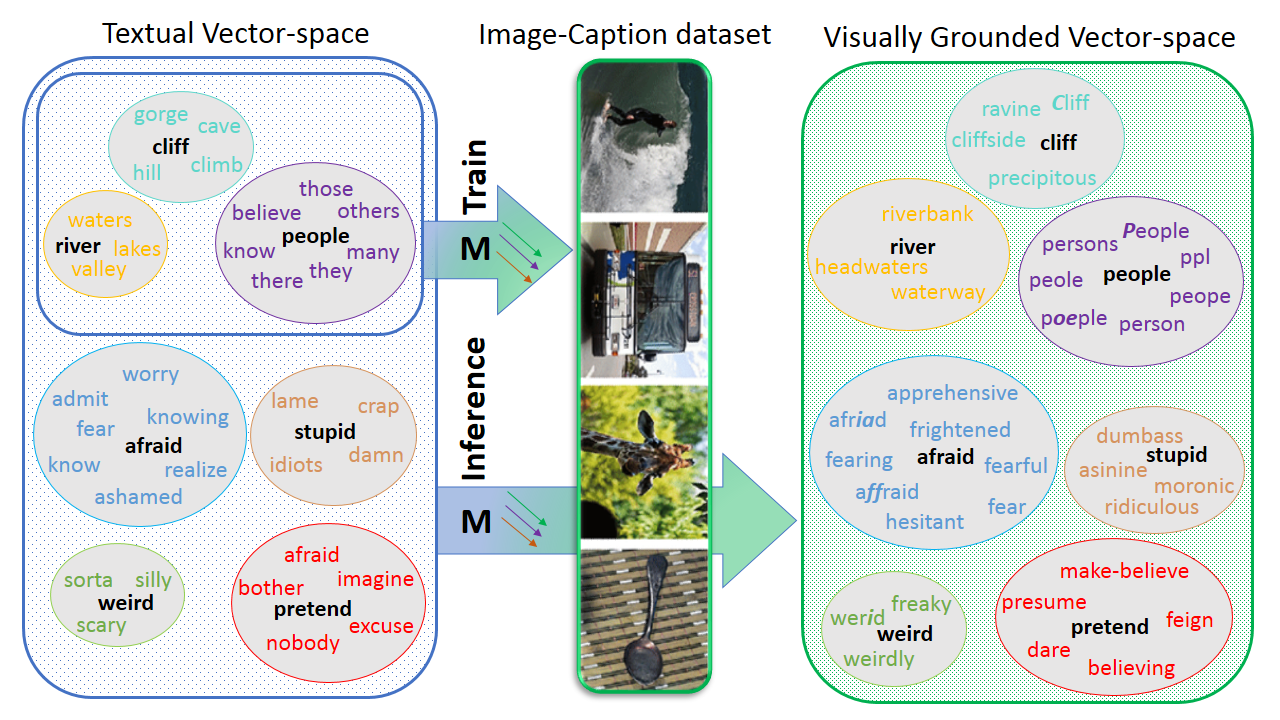}
  \caption{ \label{fig:teaser} Our model constructs visually grounded embeddings (right) from textual embeddings (left) by applying a learned alignment ($M$) trained on a subset of 10,000 words in image-caption pairs.  It then generates zero-shot grounded embeddings at the inference phase for a total of 2,000,000 words, including not only concrete words but also abstract words.  For each query word (in black), the grounded embeddings (right) retrieve more similar words compared to the purely textual embeddings (left) and alleviate the bias toward dissimilar words with high co-occurrence frequencies such as (\textit{many, people}). Out of the top 10 nearest neighbors for each query word, only the differing neighbors between the textual embeddings and the grounded embeddings are shown in the right-hand panel. } 
\end{figure}

\red{In the present work we make use of recent advances in machine learning, computer vision and NLP to propose a new method of computing multimodal embeddings via multimodal fusion \citep{baroni2016grounding}. Our approach falls into the latter category of grounding models where rather than projecting textual and visual embeddings into the same space, textual embeddings are slightly adjusted to reflect information gleaned from images (see Figure~\ref{fig:teaser}). } \red{Our model} is able to generalise to new words without a visual representation, which allows it to \red{generate grounded embeddings not only for concrete words for which images are available but also for abstract words, extending earlier work such as \citet{Johns2012PerceptualSimilarity, utsumi2022test} while making use of more recent insights from NLP. We compare our model to both ungrounded embeddings as well as embeddings based on other grounding methods and show that our model is more predictive   for responses in a range of behavioural datasets such as similarity/relatedness judgements \citep[e.g. MEN,][]{bruni2014multimodal} which have been used in previous work to evaluate word embeddings from a psycholinguistic perspective \citep{mandera2017explaining}.} Our grounded embeddings are made available to the community.\footnote{\url{https://github.com/Hazel1994/Visually_Grounded_Word_Embeddings_2}} 

\red{Our grounded embeddings allow us to} explore various questions which arise from previous work on grounding and generating distributed meaning representations in general, and which are crucial when aiming to model cognitively plausible meaning representations: \begin{enumerate}

    \item On the one hand, many studies have shown that combining visual information and textual information \red{is attractive from a theoretical point of view \citep[e.g.][]{andrews2014reconciling, lake2021word} and indeed} improves the quality of word embeddings \citep[e.g.][]{bruni2014multimodal,shahmohammadi2021learning, Lazaridou2016MultimodalInput}. On the other hand, purely textual embeddings are very successful even on tasks related to vision and spatial relations \citep{louwerse2009greographical, Abdou2021CanColor}, and purely visual embeddings do not perform well at predicting human similarity judgments \citep[e.g.][]{DeDeyne2021VisualMind}. Hence, the extent to which textual representations benefit from visual grounding, as well as the specific tasks and methods that are most effective, remains an open question. Apparently, a fine balance  has to be struck between too much and too little visual information in grounding.
    \red{A number of studies has attempted to explore this question from both a more technical, engineering perspective, but also from a cognitively motivated perspective. For instance, \citet{hill2014learning, rotaru2020constructing} found that how beneficial perceptual information is for resulting embeddings depends on the concreteness of the words: the more concrete the words are, the more they profit from perceptual information. We will explore to what extent perceptual knowledge from images is beneficial for acquiring high-quality and cognitively plausible embeddings, using a more modern grounding architecture.} \\
    
    \item Traditionally, embeddings are grounded on a single word basis \citep[e.g.][]{Gunther2020Images, kiela-bottou-2014-learning,bruni2014multimodal}. However, visual scenes are complex, and are usually best described not by single words, but rather by entire sentences. Equating complex scene structures with isolated words is not only counter-intuitive but also problematic when grounding abstract words since highly abstract words (e.g., \textit{justice}) are rarely depictable. It is known that language is vital for representing abstract concepts \citep{Borghi2017TheConcepts., dove2018language}. However, the interplay between language and perceptual experiences is still an open field. How do language and embodied experience together shape our understanding of abstract and concrete concepts? We will design various experiments to explore how language (here represented as word representations) and vision (images) should interact.

     {\item There exist multiple theories of how words are grounded in perceptual experiences \citep{paivio1971imagery,borghi2019words,howell2005model}. Nonetheless,  large scale grounding of abstract words into vision is still an open field. More specifically, the question still remains: how should abstract words be grounded in computational models on a large scale? In line with the theory of indirect grounding \citep{howell2005model,louwerse2011symbol}, we propose a large-scale grounding method\footnote{{Please note that our model is not a cognitive model. However, our findings provide substantial support for the indirect grounding theory.}} to effectively ground abstract words.} \\
 
    \item Newly proposed large-scale contextualized language models rely on enormous amounts of data \citep[e.g., BERT:][]{devlin2018bert}. While this leads to good performance, it is cognitively implausible, as humans encounter only a much smaller number of words over their lifetimes \citep{brysbaert2016many}. Our fourth question therefore relates to whether visual grounding is equally helpful when large amounts, or only small amounts, of training data are available: How much does the amount of training data influence the improvement of visual grounding on downstream tasks such as sentiment analysis? We will demonstrate that on corpora sizes closer to human-scale training data, visual grounding improves the quality of embeddings even on highly abstract tasks.
\end{enumerate}

To this end, our paper is structured as follows. Sections \ref{sec:proposed_method} and \ref{sec:implementation_details} introduce our method, which is evaluated in Section \ref{sec:evaluation}. In Sections \ref{sec:alignment_vs_fusion} and \ref{sec:gap}  we will address the first two aforementioned research questions. Furthermore, we will investigate the impact of grounding on task performance, specifically in state-of-the-art language processing models, with respect to the available training data in Sections \ref{sec:contextual_grounding} and \ref{sec:grounding_small_ds}.

\section{Visually Grounded Word Embeddings} 
\label{sec:proposed_method}

\begin{figure*}[ht]
  \centering
  \includegraphics[width=\textwidth]{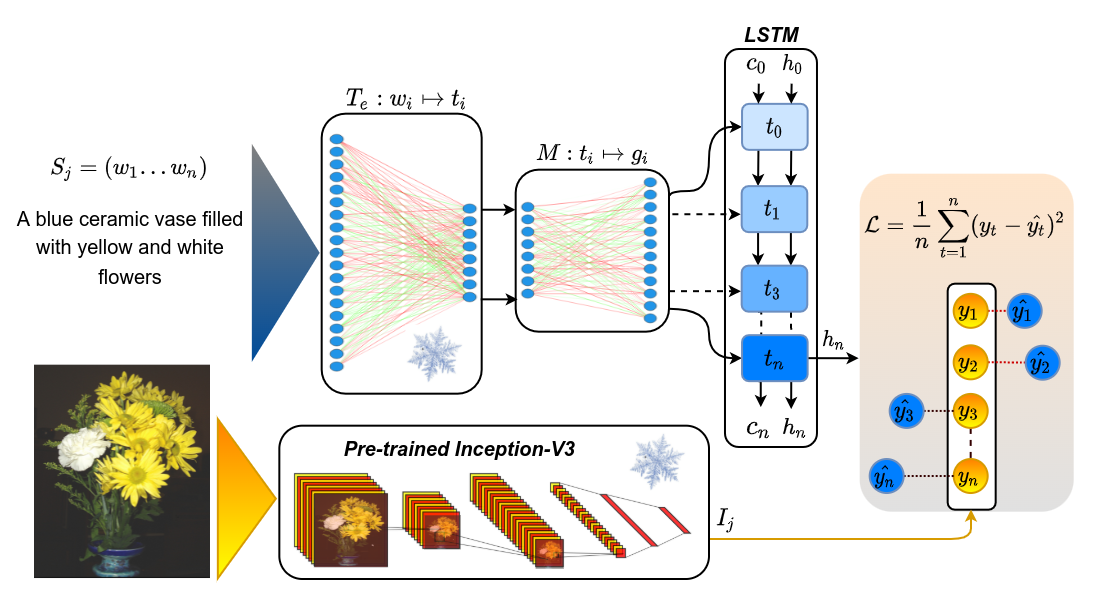}
  \caption{ \label{fig:model} Our visual grounding model encodes each caption word by word, using an LSTM, given the
task to predict the corresponding image vector. A mapping M is set up that takes textual vectors and maps
them into the grounded space. This mapping is trained on a limited number of words (those that occur in
the captions) but is then applied to all the words, after the training is completed, to generate “zero-shot” (unseen)
grounded embeddings. The snowflake icon indicates the frozen learning parameters during training.} 
\end{figure*}

In this section, we explain our visual grounding approach and how it can be used to generate visually grounded word representations from textual word embeddings. For \((S_j,I_j) \in D\), let  \(S_j=[w_1,w_2 \cdots w_n]\) be a textual caption with $n$ words describing its corresponding image with the image vector $I_j$ in the dataset $D$. {The image vector $I_j$ is obtained by feeding the image into a pre-trained convolutional neural network (CNN) model. A CNN is a family of neural networks designed for processing images with a grid-like topology that extracts local information and aggregates them through multiple layers of learnable parameters. CNNs are usually trained on a large set of images annotated by human raters to classify images into many classes (e.g., \textit{dog}, \textit{horse}, and \textit{car}). Once they are trained, they can be used to encode images into dense and meaningful numerical representations that correspond well to human intuitions \citep{bracci2019ventral, Lazaridou2017MultimodalText}}. Let $t_i\in \mathbb{R}^d$ be a textual embedding of the word $w_i$, which has been obtained by a pre-trained word embedding model $T_e:w_i\mapsto t_i$ (e.g., Fasttext). The goal is to learn a linear mapping ${M}$ to visually ground any textual word vector $t_i$ in its corresponding image vector $I_j$ and obtain the visually grounded embedding $g_i \in \mathbb{R}^c$ of the word $w_i$. {The learned mapping $M$ will linearly adjust the textual word embeddings based on the information in images.} This mapping ideally should: a) preserve the abstract knowledge from co-occurrence statistics captured by textual embeddings {trained on large textual corpora, and b) align the textual embeddings with their corresponding visual properties available in images}. This way, the grounded embeddings will benefit both concrete and abstract words \citep{shahmohammadi2021learning}. While it may seem intuitive to learn both modalities in a shared feature space, we argue that such approaches, unfortunately, are more likely to cause the grounded embeddings to lose the abstract knowledge from textual co-occurrences and therefore suffer from a bias towards concrete words as reported by \citet{park-myaeng-2017-computational}.

{It is widely acknowledged that language plays a crucial role in acquiring abstract concepts \citep{Borghi2017TheConcepts., dove2018language}. Therefore, we believe that preserving abstract knowledge during the grounding process requires individual words to be aware of the context (other words in the sentence).} The grounding process should also respect the textual vector space as any random change to textual embeddings will distort the semantic information obtained by textual statistics \citep{shahmohammadi2021learning}.  Figure~\ref{fig:model} lays out the architecture of our proposed grounding model. The grounded version of any word $w_i$ is obtained by mapping its textual embedding $t_i$ into the visually grounded space using the linear mapping $M$ as $g_i= t_i \cdot M$. In the grounded space word vectors are aligned with the images by using a one-layer Long Short-Term Memory (LSTM) network \citep{hochreiter1997long}. {The $LSTM$ network is a type of recurrent neural network that is suitable for text processing, it processes a sequence of words in a text word by word and at each step updates its internal learning parameters.} The LSTM encodes the whole sentence $S_j$ as a single vector $h_n$: \\
\begin{equation}
 h_n =  LSTM(G,c_0,h_0 \mid \theta),
\end{equation}

\noindent
where $G$ denotes the input --- all the grounded word vectors (output of $M$) --- and $\theta$ the learning parameters. It also includes a cell state $c_t$ and a hidden state $h_t$ where $t$ denotes the current time-step (the current word being processed). {At first, the network is initialized with a random hidden and cell states ($h_0$ and $c_0$)} and takes one word at each time-step (see Figure~\ref{fig:model}) and each time, for each successive word, it updates its memory by removing and adding information to the cell state. It then generates an output $h_t$ based on the current input $g_t$ and $c_t$. Both $h_t$ and $c_t$ are passed to the next time-step. We extract the output of the last time-step $h_n$ as a vector representing the whole sentence. The model is trained to match $h_n$ to the image vector $I_j$ for each particular training sample  \((S_j,I_j) \in D\). We optimize the parameters of the $LSTM$ and the mapping $M$ (denoted as $\Theta$) based on the following mean-squared-error (MSE) loss:
\begin{equation}
    \hat{\Theta} = \operatorname*{argmin}_\Theta \frac{1}{N} \sum_{t=1}^{n} (y_t - \hat{y_t})^2,
\end{equation}
where $y$ and $\hat{y}$ denote the ground truth image vector ($I_j$) and the predicted image vector ($h_n$) respectively. By applying the $LSTM$ network, the model takes into account the context in which each word occurs. Therefore, the whole sentence is mapped to the image vector. Since the model tries to predict an image vector, it will change the textual vector space such that the image vector is estimated as accurately as possible. Nonetheless,  we restrict the influence of the images on the word vectors by keeping the mapping $M$ linear. Naturally, the grounded word vectors (output of $M$) will still respect the textual vector space but they will be indirectly aligned to the image representations.\\

{After training the model on (caption, image) pairs, the mapping $M$ can be used to indirectly ground both abstract and concrete words including out-of-vocabulary words. For instance, for obtaining the visually grounded vector of the word \textit{sad}, we first fetch its textual vector $t_{sad}$ using the pre-trained textual embeddings. The grounded vector is then obtained by using the learned mapping $M$ as $g_{sad}= t_{sad} \cdot M$, where $g_{sad}$ indicates the visually grounded version of the word \textit{sad}}. In this way, a visually grounded version of the textual embeddings is created in a zero-shot manner (including unseen words) despite being exposed to only a limited number of words while training on image captions.

\section{Implementation Details}
\label{sec:implementation_details}

We used the Microsoft COCO 2017 dataset \citep{lin2014microsoft} in our experiments. Each sample of this dataset includes a single image along with 5 different human-generated captions \citep{chen2015microsoft}. The whole dataset was divided into $118k$ train and $5k$ validation samples. We set the batch size to $256$ with each batch containing $256$ image vectors (of dimension 2048) along with one of their corresponding captions. Image vectors were extracted from the penultimate layer of a pre-trained Inception-V3 CNN model \citep{szegedy2016rethinking}, based on ImageNet \citep{deng2009imagenet}. We set the dimension of the grounded embeddings (output of $M$) to $1024$, following \citet{shahmohammadi2021learning}. A one-layer $LSTM$ was  applied with $2048$ units. We removed the punctuation marks from the captions and converted all words to lowercase. Only the top $10k$ most frequent words in the captions were used and the rest were ignored. {Reducing the number of processed words is a common practice in NLP, as many words occur rarely in the training corpus and therefore make a negligible contribution to the learning process.} We trained the model for $20$ epochs ($20$ iterations on the whole dataset) with $5$ epochs tolerance early stopping, using the NAdam optimizer \citep{dozat2016incorporating} with a learning rate of $0.001$. {Early stopping is a technique to prevent a model from overfitting to the training data by stopping the training process once the model's performance on a validation dataset stops improving. In our setup, we train the model until its validation score decreases for five consecutive epochs, after which the training process is halted using early stopping.}

Both the pre-trained textual embedding $T_e$ and the Inception-V3 model are frozen --- weights are kept fixed --- during training. Two popular pre-trained textual word embeddings,  GloVe ($crawl-300d-2.2M-cased$) and Fasttext ($crawl-300d-2M-SubW$), were used to initialize the embedding $T_e$. {Therefore, we generated two sets of grounded embeddings, one from Fasttext and one from GloVe.} 


\section{Evaluation} \label{sec:evaluation}
In this section, we develop several evaluation techniques to study the behavior of visually grounded embeddings and address the initial question of how much and in what specific applications perceptual information from images contributes to the creation of high-quality and cognitively plausible embeddings.

\subsection{General Evaluation}
The question of how to appropriately evaluate word embeddings persists, despite the existence of numerous evaluation benchmarks \citep{wang2019evaluating}. However, in both psycholinguistics and NLP, humanly annotated lexical semantic similarity or relatedness datasets are commonly used to evaluate (multi-modal) embeddings \citep{mandera2017explaining, rotaru2020constructing, DeDeyne2021VisualMind,  park-myaeng-2017-computational}. Here, the task is to estimate the similarity/relatedness score of a given pair of words with the Spearman correlation as evaluation metric. { Relatedness is based on topical match which quantifies the degree to which two words are associated with each other (\textit{child}-\textit{play}). Similarity is based on taxonomic closeness which is a subset of relatedness and quantifies how alike two words are (\textit{car}-\textit{automobile})}. It is worth noting that some datasets do not distinguish between similarity and relatedness. For example, the pair (\textit{clothes}, \textit{closet}) comes with the score of $1.96$ (out of $10$) in SimLex999, but exactly the same pair receives a score of 8.00 in WordSim353, which does not distinguish between similarity and relatedness. {We assess the quality of our visually grounded word representations} using the following datasets and juxtapose the results with textual embeddings and related previous works.\\

\noindent{\textbf{MEN} \citep{bruni2014multimodal}:} {This dataset is compiled specifically for the purpose of evaluating multi-modal models. It only contains words that appear as image labels in the ESP-Game\footnote{\url{http://www.cs.cmu.edu/~biglou/resources/}} and MIRFLICKR-1M16\footnote{\url{https://press.liacs.nl/mirflickr/}} datasets. Therefore, it is suitable for multi-modal assessments. MEN consists of 3,000 word pairs with semantic relatedness ratings obtained via Amazon Mechanical Turk. For example, (\textit{sun}, \textit{sunlight}) has a MEN score of $50$ (out of $50$) but the score of (\textit{zebra}, \textit{bakery}) is $0$.
}

\noindent{\textbf{WordSim353} \citep{finkelstein2001placing}:} { This collection contains 353 word pairs annotated by 13 to 16 human judgments for each pair. The judges did not distinguish between similarity and relatedness. For instance, (\textit{computer}, \textit{keyboard}) comes with a score of $7.62$ (out of $10$).
}

\noindent{\textbf{SimLex999} \citep{hill2015simlex}:} {Unlike WordSim353, SimLex999 draws a clear distinction between similarity and relatedness as mentioned above. SimLex999 contains 999 word pairs  annotated by 500 annotators via Amazon Mechanical Turk. Both WordSim353 and SimLex999 have been used for explaining human performance in psycholinguistic tasks \citep{mandera2017explaining}.
}

\noindent{\textbf{Rare-Words} \citep[RW,][]{luong-etal-2013-better}:} { This dataset measures the performance of a word-embedding model on rare words that occur less frequently (based on Wikipedia). It contains $2034$ word pairs annotated by $10$ human judges. Examples of words in this collection are \textit{interjection} and \textit{behaviorist}.
}

\noindent{\textbf{MTurk771} \citep{halawi2012large}:} { MTurk771 consists of $771$ word pairs. The authors used WordNet\footnote{\url{https://wordnet.princeton.edu/}} to extract both related and unrelated word pairs and collected 20 human ratings for each word pair.
}

\noindent{\textbf{SimVerb3500} \citep{gerz-etal-2016-simverb}: { This dataset provides human ratings for the similarity of 3,500 \textit{verb} pairs. Providing broad coverage of verbs, this dataset offers a great resource for a better understanding of ``the complex diversity of syntactic-semantic verb behaviours'' \citep[p. 2174]{gerz-etal-2016-simverb}.
}
}

\begin{table*}[ht]
\centering
\begin{adjustbox}{width=1\textwidth}
\begin{tabular}{lccccccccc}
\hline \textbf{Model} & \textbf{RW} & \textbf{MEN} & \textbf{WSim} & \textbf{MTurk} & \textbf{SimVerb} & \textbf{SimLex} & \textbf{Mean}\\
&  &  & \textbf{353} & \textbf{771} & \textbf{3500} & \textbf{999} & \\ \hline

GloVe   & 45.5 & 80.5 & 73.8 & 71.5 & 28.3 & 40.8 & 56.7 \\
ZSG-G (ours) &   \bftab 53.2 & \bftab ***85.1 & \bftab ***78.8 & \bftab ***73.2 & \bftab ***38.5 & \bftab ***52.6 &  \bftab 63.6 \\
\hline
Fasttext     & 56.1 & 81.5 & 72.2 & \bftab ***75.1 & 37.8 & 47.1 & 61.6\\
ZSG-F (ours)  & \bftab ***57 & \bftab ***84.4 & \bftab 72.3 & 74.5 & \bftab ***39.6 & \bftab ***49.6 & \bftab 62.9\\
\hline
VGE-G &  52.6 & \bftab 85.1 & \bftab **78.9 &  \bftab ***73.4 &  37.4 & 51.8 &63.2 \\
ZSG-G (ours) & \bftab **53.2 & \bftab 85.1 & 78.8 &73.2 & \bftab ***38.5 &  ***52.6 & \bftab 63.6 \\

Cap2Both     & 48.7 & 81.9 & 71.2 & \_ &\_ & 46.7 & \\
Cap2Img      & 52.3 & 84.5 & 75.3 & \_ &\_ & 51.5 &\\
Park \& Myaeng  & \_ & 83.8 &77.5 &\_ &\_ & \bftab 58.0 & \\
P\&M\_VG.  & \_ & \_ & \_ &\_ &\_ & 15.7 & \\
Collell et al.           & \_ & 81.3 & \_ &\_ &28.6 & 41.0 & \\
\hline
\end{tabular}
\end{adjustbox}
\caption{\label{nlp_eval} Comparison of our grounded embeddings (ZSG-*) to textual embeddings and other visually grounded embedding models. Our embedings show stronger correlation with human ratings on most of the datasets. The metric is Spearman’s $\rho \times 100$. Number with stars indicate statistically significant differences ($p < 0.05\ *; p < 0.01 **; p < 0.001 ***$, t-tests) {between our grounded embeddings (ZSG-G) and textual (GloVe or Fasttext) or VGE-G embeddings.}}

\end{table*}

Table~\ref{nlp_eval} shows the evaluation results on lexical semantic benchmarks. Our zero-shot grounded embeddings are shown as ZSG-G and ZSG-F indicating the grounded versions of GloVe and Fasttext respectively. The initial segment of the table demonstrates that ZSG-G exhibits superior efficacy compared to textual GloVe across \emph{all} benchmarks. In the case of Fasttext on the other hand, improvements 
are somewhat more modest, 
probably because Fasttext takes into account sub-word information.{That is, it takes advantage of the internal structure of a word to improve vector representations. For instance, the word vector of \textit{eating} might be a combination of the \textit{eat} and \textit{ing}.  Hence, it might capture word similarity/relatedness better compared to GloVe which treats each word as a unique item.} In the lower part of the table, we compare the performance of our best model (ZSG-G) with related visually grounded embedding models. For a fair comparison, we limit our list to those who adopted pre-trained word embeddings. 
\cite{shahmohammadi2021learning} (shown as VGE-G in the table) proposed a similar grounding approach to ours where {they train a linear mapping to transfer from textual word representations to visually grounded representations. However, the main difference with our approach is the training scheme of the mapping. While we only train using a single task (predicting the associated image vector given its caption), multi-task training with 3 different tasks is adopted in their approach. In their setup, the model generates the corresponding caption word by word for a given image vector in both forward and backward directions. Furthermore, the model receives pairs of captions and images as inputs and learns to discriminate between matching and non-matching pairs.} While inspired by their method, our approach is simpler, requires less computational power, and performs slightly better on the same set of benchmarks. 
\citet{kiela-etal-2018-learning} also { proposed a visual grounding approach for pre-trained textual word representations (GloVe), by using the same image database as ours. Similar to \citet{shahmohammadi2021learning} their approach is based on multi-task training where the following tasks have been proposed:} Cap2Img: predicting the image vector from its caption; Cap2Cap: generating an alternative caption of the same image; Cap2Both: training by Cap2Cap and Cap2Img simultaneously. Our approach, despite its simplicity, captures the semantic relationships of words much better compared to Cap2Both and Cap2Img. {Next, we compared our results with polymodal embeddings by \citet{park-myaeng-2017-computational}. In this approach, the meaning of each word is derived from six different types of distinct embeddings including linear context, syntactic context, visual perception, cognition, emotion, and sentiments based on the human cognitive model proposed by \citet{maruish2013clinical}.} Even though their approach uses more resources including two pre-trained embeddings (Word2Vec, GloVe) and incorporating other modalities, ours is still superior on MEN and WSim353, albeit worse on Simlex999. {The large performance gap observed for SimLex999 may be attributed to the multi-modality training of the model conducted by \citet{park-myaeng-2017-computational}. Employing solely their visually grounded embeddings (P\&M\_VG) results in low-quality word vectors, further confirming that their visually grounded embeddings do not benefit abstract words \citep{park-myaeng-2017-computational}. }

{
For further consolidation, we calculated the t-test\footnote{ \href{https://docs.scipy.org/doc/scipy/reference/generated/scipy.stats.ttest_ind.html}{}} \citep{student1908probable} between the predictions of textual and grounded embeddings for both GloVe and FastText and compared the results of our grounded GloVe (ZSG-G) with the previous VGE-G by \citet{shahmohammadi2021learning} (denoted as *, **, or *** in Table~\ref{nlp_eval}). All the improvements over the textual embeddings were found to be statistically significant with the exception of \textit{RW} dataset using GloVe. The differences in performance between our embeddings and VGE-G were found to be significant across all the benchmarks.}


{In summary, our approach while trained on a limited number of words available in image captions, creates visually informed word representations, even for unseen words, that are more aligned with human judgment across a wide range of human-rated word similarity and relatedness tasks.}






\subsection{Fine-Grained Evaluation on Concrete and Abstract Words}
In linguistics, concrete words\footnote{We assume individual words, as they are realized in English writing conventions, are the verbal expression of lexical concepts in language, and thus the terms ``word'' and ``concept'' are used interchangeably in this section.} refer to physically real and perceptible entities such as \textit{tree}, \textit{ball}, or \textit{Chris}, whereas abstract words have references that are not readily perceptible to the senses, and are more complex and variable in meaning, including mental states (e.g., \textit{happiness}), events (e.g., \textit{encounter}), conditions (e.g., \textit{totalitarianism}), relations (e.g., \textit{brotherhood}) and so forth \citep{ApaDictionary,Borghi:Binkofski:2014,barsalou2018moving, davis2020situational}. 
Concreteness and abstractness are not binary properties of words \citep{Wiemer-Hastings2001}. Words become increasingly abstract as they are more separated from physical entities and more linked to mental states \citep{Barsalou2003a}.
Word concreteness indicates the degree to which a word denotes a perceptible entity and is measured on a numerical scale by subject ratings \citep{Brysbaert2014}. For example, the word \textit{pancake} is ranked high on the scale as it is associated with many sensory properties such as smell, taste, shape, color, etc.

Extensive evidence from behavioral experiments suggests that there is an advantage in cognitive processing of words for concrete over abstract words---often referred to as the ``concreteness effect''. It has been shown that concrete words, compared to abstract words, are processed faster in isolation \citep{Schwanenflugel:Shoben:1983} and non-supportive contexts \citep{Schwanenflugel:Stowe:1989}, are remembered better in paired associative learning \citep{Paivio:1965} and free recall tasks \citep{Schwanenflugel:Akin:Luh:1992}, and are learned faster \citep{Mestres-Misse2014}. Evidence has been put forward for this distinction in the brain. Case reports of patients with brain damage demonstrate differential impairments with regard to abstract and concrete concepts \citep{Breedin1994, tyler1995abstract, Warrington1975}. Neuroimaging studies provide evidence for overlapping but distinct brain areas engaged in the processing of abstract and concrete concepts \citep[see][for a review]{Montefinese2019}.

To investigate the influence of grounding on abstract and concrete words, we leverage the SimLex999 dataset. It divides its words into different categories including  adjectives, nouns, verbs, concreteness quartiles (from 1 to 4 increasing the degree
of concreteness), and `hard' sections. {The `hard' section includes the 333 most associated word pairs in the University of South Florida Free Association Database (USF) \citep{nelson2004university}. This subset of SimLex999 is reported to be the hardest for semantic models to capture because the noise from the high association makes it hard to distinguish between similarity and relatedness \citep{hill2015simlex}. Examples of this category are \textit{happy-cheerful} and \textit{weird-strange}.} Table~{\ref{tab:simlex}} shows our fine-grained evaluation on SimLex999. {We compared our fine-grained results with that of Picturebook, another kind of visually grounded embeddings \citep{kiros-etal-2018-illustrative}. For each word, Picturebook retrieves the top-k images using  image search. The retrieved images are then passed through a CNN trained with a semantic ranking objective with 100+ million images \citep{wang2014learning}. The grounded embedding of each word is computed based on a combination of image vectors and the pre-trained GloVe embedding of that word.} Our best model (ZSG-G) captures semantic relationships much better compared to other visually grounded embeddings and generalizes across different word types. For example, it not only demonstrates a more pronounced association with highly concrete (Conc-q4) words by a margin of $19.2$ percentage points, but also with highly abstract words (Conc-q1) by a margin of $11.3$ percentage points compared to the textual GloVe vectors. In contrast, PictureBook \citep{kiros-etal-2018-illustrative}, for example, highly benefits the more concrete words but adversely affects the more abstract category even when combined with GloVe embeddings. In comparison with VGE-G by \cite{shahmohammadi2021learning}, our model again achieves better results while being much simpler and less computationally expensive.\\



\begin{table*}[ht]
\centering
\begin{adjustbox}{width=1\textwidth}
\begin{tabular}{lc|cccccccc}
\hline \textbf{Model} & \textbf{All} & \textbf{Adjs} & \textbf{Nouns} & \textbf{Verbs} & \textbf{Conc-q1} & \textbf{Conc-q2} & \textbf{Conc-q3} & \textbf{Conc-q4} & \textbf{Hard}\\ \hline

GloVe   & 40.8 & 62.2 & 42.8 & 19.6 & 43.3 & 41.6 & 42.3 & 40.2 & 27.2 \\
VGE-G  &  51.8 &  72.1 & 52.0   & \bftab 35   &  53.1 &\bftab 54.8 & 47.4 & 56.8 &  38.3\\
ZSG-G (ours)  & \bftab 52.6 & \bftab 73.8 & \bftab 53.1   & 34.6   & \bftab 54.6 & 53.9 & 48.1 & 59.2 & \bftab 39.3\\

Picturebook &37.3& 11.7 & 48.2   & 17.3   & 14.4 &27.5 &  46.2 & \bftab 60.7 & 28.8\\
Picturebook+GloVe &45.5& 46.2 &  52.1   & 22.8   & 36.7 &41.7 &  \bftab 50.4 &  57.3 & 32.5\\
\hline
\end{tabular}
\end{adjustbox}
\caption{\label{tab:simlex} SimLex999 (Spearman's $\rho \times 100$) results. 
Conc-q1 and Conc-q4 indicate the most abstract and concrete words respectively. Our model (ZSG-G) demonstrates stronger associations with human annotators' similarity ratings on multiple categories.}
\end{table*}

{We further extended the analysis of abstract and concrete words by using all the word similarity/relatedness datasets. For this aim, we first combined all the datasets (see Section~\ref{sec:evaluation}) after normalizing the score of each dataset. That is, we transformed the scores to be in the range of $[0, 1]$ as follows:
$x_{in} = \dfrac{x_i - min}{max - min} $, 
where $x_{in}$ and $x_i$ indicate the new score and the original score of the $i$th word pair respectively. $max$ and $min$ denote the maximum and minimum scores within the given dataset. After normalizing and combining all the benchmarks we obtained 10657 word pairs. We then ranked all the word pairs based on a concreteness rating dataset compiled by \cite{Brysbaert2014}. This dataset contains  37k words and 3k two-word phrases rated by over 4,000 subjects using the Amazon Mechanical Turk (MTurk) crowdsourcing platform. We denote this dataset as MTurk40k. We took the intersection between MTurk40k and our combined dataset which resulted in 8936 word pairs with both similarly/relatedness and concreteness scores. We refer to this dataset as \textit{WCR} (word concreteness rating) for simplicity. The concreteness score of a word pair was obtained by taking the average scores of its constituent words. Examples of highly abstract and concrete word pairs from \textit{WCR} are \textit{(belief, purpose)} and \textit{(apple, lemon)} respectively. Having access to a large set of word pairs with concreteness scores, we can more thoroughly assess the behavior of visual grounding on abstract and concrete words. To accomplish this, we devised a new experiment that draws upon the \textit{WCR} dataset.\\
\noindent{\textbf{Concreteness vs Abstractness:}} We computed a similarity score between each pair of the WCR dataset by applying the cosine similarity to the corresponding word vectors and used the Spearman correlation as the evaluation metric. We evaluated both the textual (GloVe) and visually grounded embeddings on four distinct subsets of the WCR with different concreteness scores. Concreteness subsets are obtained by the following steps.\\}
\begin{figure*}[ht]
  \centering
  \includegraphics[width=.8\textwidth]{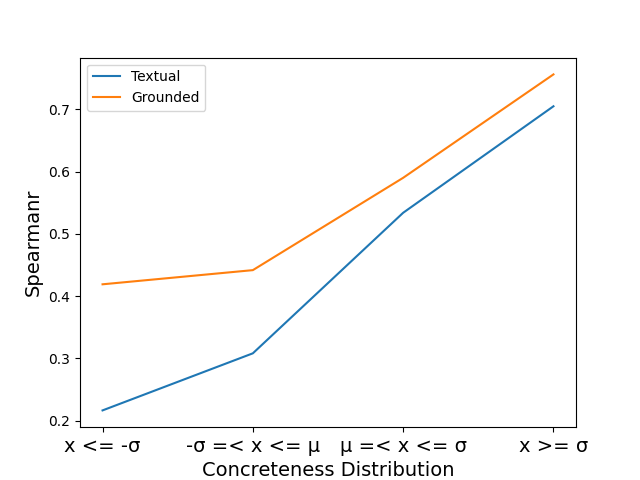}
  \caption{ \label{fig:conc_vs_abst}Comparision between textual and grounded embeddings of word pairs with different concreteness scores. Visually grounded embeddings highly benefit abstract concepts. $x >= \sigma$ and $x <= -\sigma$ indicate highly concrete and highly abstract words accordingly.} 
\end{figure*}
{
\begin{enumerate}
    \item To account for variations in concreteness scores, a standardization procedure is applied whereby scores are transformed into a standard normal distribution. Specifically, this involves subtracting the mean from all scores and dividing by their standard deviation, resulting in a standardized score $x_{is}$ for the $i$th word pair, expressed as $x_{is}:\dfrac{x_{in} - \mu}{\sigma}$.
    \item After standardization, the distribution is partitioned into four segments based on the standard deviation and mean values, namely $[-\sigma, \mu, \sigma]$. The placement of word pairs within these segments allows for the differentiation of concrete and abstract word pairs. Specifically, pairs with higher concreteness scores are more likely to fall on the right side of the distribution ($x > \mu$), while those with lower scores are more likely to be located on the left side of the distribution ($x < -\mu$).
\end{enumerate}
\begin{figure}
     \centering
     \begin{subfigure}[b]{0.45\textwidth}
         \centering
         \includegraphics[width=\textwidth]{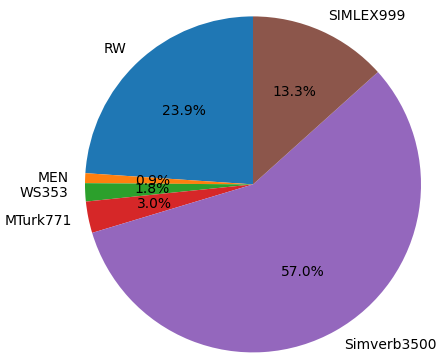}
         \caption{abstract subset ($x < - \sigma$)}
         \label{fig: abst_prop}
     \end{subfigure}
     \hfill
     \begin{subfigure}[b]{0.45\textwidth}
         \centering
         \includegraphics[width=\textwidth]{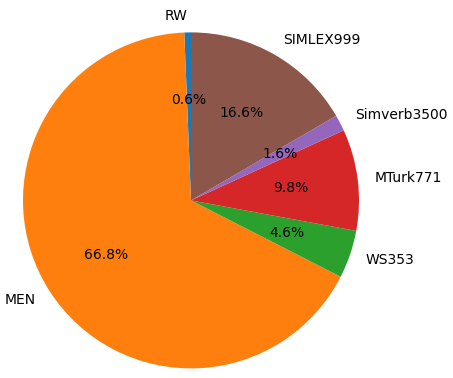}
         \caption{concrete subset ($x > \sigma$)}
         \label{fig:conc_prop}
     \end{subfigure}
     \hfill
    \caption{Dataset proportions for the highly abstract and highly concrete subsets of word pairs.}
    \label{fig:data_proportion}
\end{figure}
Results are shown in Figure~\ref{fig:conc_vs_abst}. Visual grounding leads to improved quality of textual embeddings regardless of the degree of concreteness. While the embeddings capture the meanings of concrete words more accurately in general, the improvement is more significant for highly abstract words ($x <= -\sigma$). To investigate the potential cause of higher improvements for abstract words, we plotted the datasets' proportions of highly concrete words and highly abstract words in Figure~\ref{fig:data_proportion}. Highly abstract word pairs are dominated by the \textit{SimVerb3500} dataset, which seems to be the hardest for the textual embeddings to model (see Table~\ref{nlp_eval}). Highly concrete word pairs on the other hand mostly originate from the \textit{MEN} benchmark, perhaps unsurprisingly,  as it was compiled from image labels. The textual embeddings perform the best on this benchmark. Our finding is in line with previous works indicating that the meaning of concrete words is more stable and reliable compared to abstract words across different textual word embeddings \citep{pierrejean2019investigating}. \\}

\noindent{\textbf{Concreteness Separation:}} Thus far, our findings demonstrate that the use of visual grounding leads to an improvement in the quality of embeddings for both concrete and abstract words. It is reasonable to assume that this is due to the grounding process creating a clearer separation between these two types of words. We carried out the following experiments to see whether this hypothesis holds true. We conducted training and assessment of two regression models by employing 10-fold cross-validation on the MTurk40k dataset, which is a concreteness rating dataset assembled by \citet{Brysbaert2014}. {The models utilized in this experiment included a straightforward linear regression and a multi-layer perceptron (MLP). The architecture of the MLP incorporated two hidden layers with 512 and 100 neurons, respectively. The models were given word representations as input and trained to predict the standardized concreteness scores.\footnote{https://scikit-learn.org/stable/modules/generated/sklearn.preprocessing.StandardScaler.html}. Additionally, batch normalization \citep{ioffe2015batch} and dropout \citep{srivastava2014dropout} techniques were integrated into the MLP model for better generalization. Dropout is a regularization technique to prevent overfitting by randomly dropping out (setting to zero) some neurons during training. Batch normalization improves the stability and speed of training by normalizing the inputs to each layer.} Reported in Table~\ref{tab:conc}, the difference between GloVe and our grounded embeddings (ZSG-G) is very subtle. This shows that visual grounding, as implemented in our model, does not necessarily cause stronger discrimination between concrete and abstract words.\\

\begin{table*}[ht]
\centering
\begin{tabular}{l|cc}
\hline \textbf{Model} & \textbf{GloVe 10-fold-score} & \textbf{ZSG-G 10-fold-score} \\ \hline

Linear regression   & 84.90 & 84.70 \\
Multi-layer-perceptron & 88.86 & 88.24\\
\hline
\end{tabular}
\caption{\label{tab:conc} Mean Spearman's correlation coefficient $\times 100$ on MTurk40k using 10-fold-CV. Visually grounded embeddings (ZSG-G) do not seem to separate concrete and abstract words better in comparison to textual embeddings (GloVe).}
\end{table*}

\noindent{\textbf{Nearest Neighbors:}} For further exploration, we juxtaposed a sample of differing nearest neighbors of our best embeddings (ZSG-G) with its purely textual version (GloVe). Figure~\ref{fig:teaser} shows the results for two random samples of highly abstract and highly concrete words in SimLex999. While GloVe retrieves related words (shown on the left), our grounding shifts the focus toward similarity and retrieves highly similar words for both concrete and abstract queries (shown on the right). We can observe that GloVe suffers from a bias toward the dissimilar words that frequently co-occur such as (many, people) and (sorta, weird). Our embeddings, on the other hand, alleviate this bias by {creating more refined clusters of words. Even though our alignment is trained with mostly concrete words, the resulting vector space also benefits abstract words. In other words, abstract words are grounded indirectly via a learned mapping trained with concrete words. These findings align with the perspective of indirect grounding, which posits that concrete words are directly grounded while abstract words are indirectly grounded through language \citep{howell2005model,louwerse2011symbol, hoffman2018concepts}. Indirect grounding of abstract words has recently shown promising results in predicting abstract concepts using distributional semantic models \citep{utsumi2022test}.
} Moreover, different typos of the same word such as `peope' and `poeple' (for people) occur with different frequencies in different contexts. Therefore, they are gradually pulled apart. Our model, however, puts them back into the same vicinity of space by applying the learnt alignment.

\section{Alignment vs Fusion}\label{sec:alignment_vs_fusion}
{In this and the subsequent section, we will conduct new experiments that manipulate the relationship between language and vision. These experiments will contribute to gaining deeper insight into the second question raised: how might language and embodied experiences work together to shape our comprehension of words?} As the first step, various scenarios in which visual information could enhance textual word vectors are explored. In other words, we are interested to see whether increasing the influence of images on word vectors results in better grounded word vectors. For this aim, we train our model (ZSG-G) with different activation functions for the mapping $M$. Using a non-linear activation function such as ReLU and Leaky-ReLU \citep{xu2015empirical} and adding more non-linear layers will allow the model to drastically deform the textual vector-space beyond linear transformations, increasing the influence of images on grounded word vectors. Table~\ref{tab:act_effect} shows the results with different numbers of layers and non-linear activation functions. We measure similarity and relatedness by evaluating on MTurk771 and SimLex999, as they are compiled for similarity and relatedness respectively. Leveraging from different categories in SimLex999, we also evaluate on highly abstract and highly concrete words. Furthermore, for each case, we evaluate the obtained word vectors on all of the available datasets mentioned in Table~\ref{nlp_eval}. As shown in Table~\ref{tab:act_effect}, we observe a consistent pattern of losing abstractness and gaining concreteness when non-linear transformations are used. This is to be expected, since word vectors are morphing into image vectors and hence gain concrete properties. Employing two consecutive Leaky-ReLU is a prominent example of this case. Results on similarity and relatedness show that visual grounding shifts the focus toward similarity (see also Figure~\ref{fig:teaser}). However, both similarity and relatedness are improved compared to textual embeddings by using a linear transformation, which helps benefiting from vision while keeping the textual information preserved. Overall, the best results on all the datasets are achieved by the linear mapping. { This suggests that while visual information is beneficial for enhancing textual embeddings, giving too much emphasis to vision and neglecting language is not the optimal approach. These findings support previous evidence from case studies, as well as behavioral and neural studies, which suggest that abstract and concrete words are processed differently and involve distinct but overlapping brain regions \citep[see][for reviews]{Montefinese2019, Mkrtychian2019}. Therefore, it is crucial to strike a balance between concreteness and abstractness, which are represented in our experiments by visual properties of images and statistics of textual corpora respectively. Language seems to benefit from vision the most when it is aligned/informed with vision as opposed to being completely fused together.}
As the first step, various scenarios in which visual information could enhance textual word vectors are explored. In other words, we are interested to see whether increasing the influence of images on word vectors results in better grounded word vectors. For this aim, we train our model (ZSG-G) with different activation functions for the mapping $M$. Using a non-linear activation function such as ReLU and Leaky-ReLU \citep{xu2015empirical} and adding more non-linear layers will allow the model to drastically deform the textual vector-space beyond linear transformations, increasing the influence of images on grounded word vectors. Table~\ref{tab:act_effect} shows the results with different numbers of layers and non-linear activation functions. We measure similarity and relatedness by evaluating on MTurk771 and SimLex999, as they are compiled for similarity and relatedness respectively. Leveraging from different categories in SimLex999, we also evaluate on highly abstract and highly concrete words. Furthermore, for each case, we evaluate the obtained word vectors on all of the available datasets mentioned in Table~\ref{nlp_eval}. As shown in Table~\ref{tab:act_effect}, we observe a consistent pattern of losing abstractness and gaining concreteness when non-linear transformations are used. This is to be expected, since word vectors are morphing into image vectors and hence gain concrete properties. Employing two consecutive Leaky-ReLU is a prominent example of this case. Results on similarity and relatedness show that visual grounding shifts the focus toward similarity (see also Figure~\ref{fig:teaser}). However, both similarity and relatedness are improved compared to textual embeddings by using a linear transformation, which helps benefiting from vision while keeping the textual information preserved. Overall, the best results on all the datasets are achieved by the linear mapping. { This suggests that while visual information is beneficial for enhancing textual embeddings, giving too much emphasis to vision and neglecting language is not the optimal approach. These findings support previous evidence from case studies, as well as behavioral and neural studies, which suggest that abstract and concrete words are processed differently and involve distinct but overlapping brain regions \citep[see][for reviews]{Montefinese2019, Mkrtychian2019}. Therefore, it is crucial to strike a balance between concreteness and abstractness, which are represented in our experiments by visual properties of images and statistics of textual corpora respectively. Language seems to benefit from vision the most when it is aligned/informed with vision as opposed to being completely fused together.}

\begin{table*}[ht]
\centering
\begin{adjustbox}{width=1\textwidth}
\begin{tabular}{lccccccc}
\hline \textbf{Type-Act.(No. of Layers)} & \textbf{Relatedness} & \textbf{Similarity} & \textbf{Abstract} & \textbf{Concrete} & \textbf{All}\\
\hline
Textual GloVe   & 71.5 & 43.3 & 43.3 & 40.2 & 56.7 \\

Grounded-Linear(1)   & \bftab 73.2 &  52.6 & \bftab 54.6 & 59.2 & \bftab 63.6\\
Grounded-ReLU(1)   & 69.2 & 50.1 & 49.4 & 60.5 & 59.7\\
Grounded-Leaky-ReLU(1) & 73.0 & \bftab 53.9 & 52.8 & 61.7 & 63.0\\
Grounded-Leaky-ReLU(2) & 71.3 & 52.4 & 49.6 & \bftab 64.6 & 61.7\\
\hline
\end{tabular}
\end{adjustbox}
\caption{\label{tab:act_effect} The impact of various activation functions and the number of layers used for the mapping $M$. on-linear transformations led to a reduction in abstract knowledge and an increase in concreteness. The term ``All'' refers to the average score across all datasets listed in Table~\ref{nlp_eval}.}

\end{table*}

\section{Bridging the Gap Between Language and Vision} \label{sec:gap}

While our model is relatively simple compared to many others \citep{shahmohammadi2021learning,kiros-etal-2018-illustrative,kiela-etal-2018-learning}, there are alternative approaches that use even simpler methods to integrate language with vision \citep{collell2017imagined,Gunther2020Images,hasegawa-etal-2017-incorporating}. This raises the question of how to properly fill the gap between language and vision. We therefore { investigated different ways in which the part of our model that bridges this gap can be engineered, and evaluated how well these alternative implementations perform.}
We constructed the following scenarios. In all the scenarios, similar as before, after the training, we use the trained mapping $M$ to map all the textual embeddings into the grounded space to obtain grounded embeddings.\\   

\noindent \textbf{Word-Level (WL):} For each training (caption, image vector) pair $(S_j, I_j) \in D$, we remove the stop words in caption $S_j$ and train a linear mapping $M$ from each word to its  corresponding image vector $I_j$. For instance, the caption \textit{`there is a dog on the floor'} would be converted into \textit{`dog floor'}. Then, the textual embeddings of both \textit{dog} and \textit{floor} are mapped to their corresponding image one by one using only the mapping $M$. Similar to \citet{Gunther2020Images}, we employed PCA \citep{pearson1901liii} to match the dimensions of the image vectors (2048) to the output of the mapping $M$ (1024).\\

\noindent \textbf{Bag-of-Words (BoW):} For each training (caption, image vector) pair $(S_j, I_j) \in D$, after mapping all the words in $S_j$ into the grounded space using a linear mapping here denoted again as $M$, we average them to obtain the BoW sentence representation. The BoW vector is then mapped into the image vector $I_j$ using a hidden layer with $Tanh$ activation function. This approach represents a more sophisticated method than the 'Word-Level' model, as it utilizes all words in the captions and incorporates a non-linear transformation, potentially leading to improved performance.\\

\noindent \textbf{GRU:} This set-up is very similar to our proposed model (see Section~\ref{sec:proposed_method}), and differs in that a single-layer GRU \citep{cho2014learning} is used instead of an LSTM. A GRU is less complex compared to an LSTM and contains only a hidden-state as opposed to the LSTM, which is equipped with both a cell-state and a hidden-state.\\

\noindent \textbf{LSTM:} This refers to the model proposed in Section~\ref{sec:proposed_method}. \\

\noindent \textbf{Transformer-Encoder (TE):} Attention-based sequence encoders introduced in \cite{vaswani2017attention} are currently used in state-of-the-art contextualized language models \citep{lan2019albert, devlin2018bert} {and are applied to  complex downstream NLP tasks.  We are interested in whether the utilization of cutting-edge NLP techniques can enhance the capacity to capture human-rated word similarity and relatedness. These encoders generate contextualized embeddings based on the learnable associations between words, allowing for the disambiguation of polysemous words in different contexts.} For instance, the word \textit{`clip'} has different senses in \textit{`I clip my nails'} and \textit{`I saw a video clip'}. To distinguish between these senses, contextualized representations of \textit{`clip'} are therefore computed that are informed by its associations with the words in a given context. For our experiments, we pass the textual embeddings of each caption through the mapping $M$ as before. Then we train a different number of encoders on top of $M$. {That is, the embeddings are passed through multiple transformer encoders simultaneously. The output of the encoders is the contextualized representation of the given caption} which is then projected to the image vector through a linear layer. We constructed the transformer encoders with $1024$ hidden size, $16$ attention heads and used NAdam with the learning of $0.0001$ for training. For a comprehensive understanding of transformer architecture, we highly recommend referring to the seminal work by \citet{vaswani2017attention}.

The results of each model configuration are reported in Table~\ref{tab:fus_vs_algn}. Notably, the Word-Level mapping fails to preserve a sufficient amount of textual information, resulting in embeddings that are significantly distorted when compared to text-only embeddings.  As a consequence, these embeddings demonstrate inferior performance across all datasets.  We note here that a single image is very rich in information and often is not well-described by a single word. {Furthermore, the relationship between language and vision is not always linear or straightforward. For instance, many highly concrete nouns and adjectives such as \textit{apple} and \textit{red} could be easily coupled with their visual representations. In contrast, more abstract linguistic categories such as prepositions and conceptual words establish their link to visual experiences through intricate (not necessarily linear) statistical patterns embedded within language. }

While the BoW model does offer some improvement over the text-only GloVe approach on certain datasets, its overall performance is relatively comparable. However, it is worth noting that the BoW model demonstrates significant enhancement on the SimLex999 dataset, which evaluates word similarity rather than relatedness. Conversely, its performance is weaker on the MTurk771 dataset, which focuses on relatedness. The potential reason for these fluctuations in performance is that the BoW representations do not account for word order and, consequently, lose the temporal statistics of how related words co-occur within their context \citep[see][for embeddings jointly representing word meaning and word order]{jones2007representing}.
The utilization of recurrent neural networks (specifically, GRU and LSTM models) results in significantly improved performance. Of these two models, the LSTM outperforms the GRU, which is unsurprising given its ability to effectively capture long-distance dependencies between words and encode the entirety of a sentence.

However, training with a single transformer encoder fails to produce better quality embeddings, perhaps unsurprisingly as these encoders are usually stacked on top of each other to achieve the desired outcome \citep{vaswani2017attention}. We therefore also tested models with two and three layers of TE. While using a two-layer TE demonstrated improved performance, we did not observe any further improvement with additional layers beyond that. We also employed multiple layers of LSTM and found that a single-layer LSTM produces the most favorable outcomes. While adding more layers typically results in a more robust model, we contend that as the network grows deeper, there is a decreased amount of visual knowledge that can be easily conveyed back to the mapping $M$. In other words, the visual knowledge becomes distributed across various layers, making it arduous to distill the information down into a single layer. Recall that after the training we only use the mapping $M$ to obtain visually grounded representations. Consequently, a network that effectively condenses information within $M$ while accurately predicting image vectors is highly desirable. In our experiments, we found that a single-layer LSTM strikes the ideal balance between the degree of dependence on $M$ and producing high-quality image vectors.\\


{In summary, our experiments in the last two sections aimed to apply computational models to shed light on the question of how language and embodied experiences (here crudely represented as images) might interact to shape our comprehension of words. In our experiments}, a linear transformation in isolation is not adequate for establishing a strong connection between vision and language. In order to obtain high-quality visually grounded embeddings, it is imperative to incorporate a non-linear transformation. Furthermore, it is essential to carefully calibrate the semantic space of the textual embeddings to accurately capture the perceptual knowledge present in images. Allowing too much influence from the visual modality may lead to distortion of the textual embeddings, emphasizing the importance of striking a delicate balance between the two modalities. {This finding suggests that also the human mind integrates information from vision in its semantic system, but that this system is not dominated by visual similarities. It is worth noting that philosophers such as Kant, Husserl, and Merseau-Ponty have pointed out that we do not perceive the world as it truly is, our perceptions are shaped by our senses, the constraints imposed by the world on our survival, and our cultures \citep[see, e.g.,][]{kant1999critique, Husserl:1913,merleau2013phenomenology}. A very similar point was made more recently from the perspective of the cognition of vision by \citet{hoffman2019case}. The way in which we implement visual grounding --- constraining the extent to which vision can change embeddings from human texts --- does justice, however crude, to this fundamental insight.}
\begin{table*}[ht]
\centering
\begin{adjustbox}{width=1\textwidth}
\begin{tabular}{lccccccccc}
\hline \textbf{Model} & \textbf{RW} & \textbf{MEN} & \textbf{WSim} & \textbf{MTurk} & \textbf{SimVerb} & \textbf{SimLex} & \textbf{Mean}\\
&  &  & \textbf{353} & \textbf{771} & \textbf{3500} & \textbf{999} & \\ \hline

GloVe    & 45.5 & 80.5 & 73.8 & 71.5 & 28.3 & 40.8 & 56.7 \\
WL       & 27.7 & 49.7 & 34.2 & 31.7 & 7.10 & 1.50 & 25.3  \\
BoW     & 46.5 & 75.2 & 73.8 & 60.1 & 33.8 & 46.0 & 55.9   \\
GRU      & 51.2 & 83.0 & 75.1 & 71.3 & 36.9 & 48.3 & 60.1   \\
LSTM &   \bftab 53.4 & \bftab 85.1 & \bftab 78.8 & \bftab 73.2 & \bftab 38.5 & \bftab 52.6 &  \bftab 63.6 \\
1-layer-TE      & 44.0 & 77.4 & 62.9 & 67.0 & 25.5 & 37.5 & 52.9 \\
2-layer-TE      & 50.1 & 82.6 & 75.3 & 72.2 & 32.4 & 45.6 & 59.7 \\
3-layer-TE      & 50.0 & 82.0 & 72.7 & 72.2 & 33.0 & 46.8 & 59.4 \\
\hline

\end{tabular}
\end{adjustbox}
\caption{\label{tab:fus_vs_algn} Evaluation of various textual encoders reveals a consistent improvement in performance from the most simplistic approach (WL) to the utilization of an LSTM model. However, In light of our experimental results, it appears that transformer-encoders may not be particularly well-suited for generating visually grounded word embeddings.}

\end{table*}

\begin{figure*}[ht]
  \includegraphics[width=\textwidth]{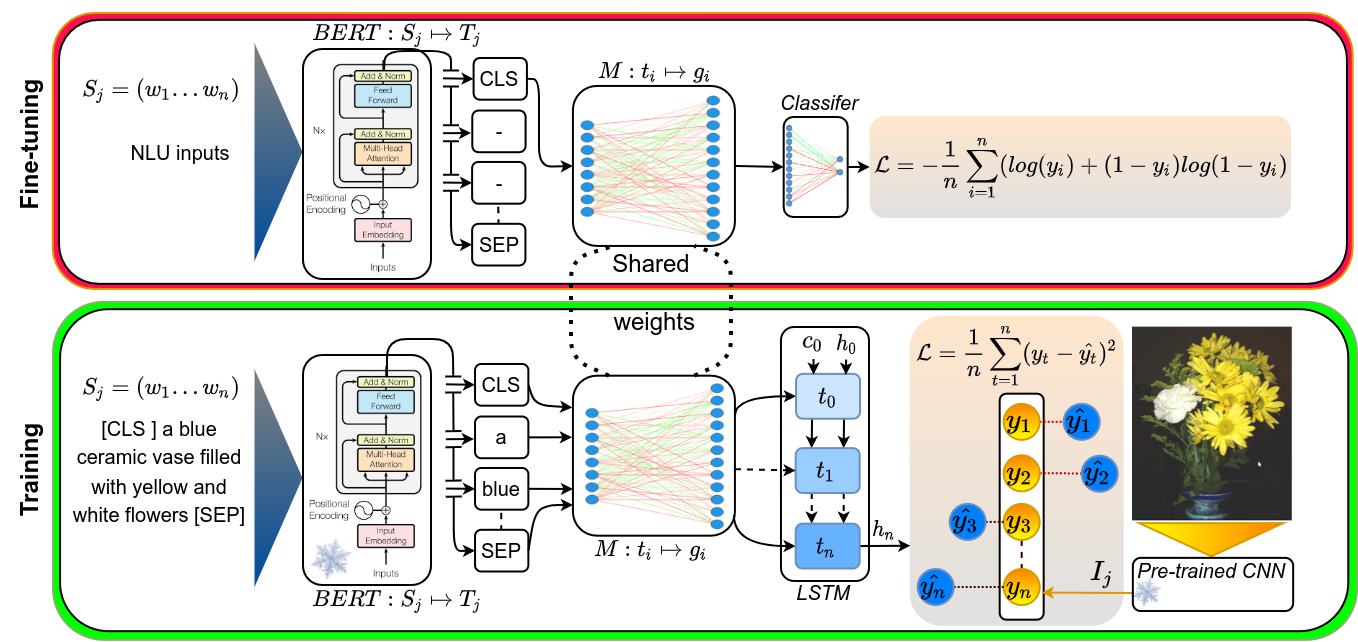}
  \caption{ \label{fig:bert}We construct a visually grounded version of BERT using image-caption pairs. In the training phase, the frozen pre-trained BERT encodes the caption, and an alignment $M$ followed by an LSTM layer on top of BERT is trained to predict the corresponding image vector. In the fine-tuning phase, the learned alignment $M$ is attached on top of BERT followed by a classifier. This alignment ensures that the BERT representations are guided by the learned visual alignment during fine-tuning.} 
\end{figure*}

\section{Contextualized Visual Grounding}\label{sec:contextual_grounding}

While we successfully showed the benefit of visual grounding for word embeddings on a wide range of intrinsic tasks, { it remains a topic of debate as to whether visual grounding provides benefits for state-of-the-art NLP models on sentence-level language tasks \citep{Yun2021DoesVP,iki2021effect,tan-bansal-2020-vokenization}. While some recent approaches have reported minor improvements through the use of visually grounded models \citep{sileo2021visual}, there is a growing consensus that these models, such as VL-BERT \citep{su2019vl}, do not provide significant benefits for language tasks. In fact, there is concern that these models may distort the linguistic knowledge acquired from textual corpora and hinder their effectiveness for natural language understanding tasks} \citep{tan-bansal-2020-vokenization, Yun2021DoesVP} {and modeling abstract concepts \citep{pezzelle2021word}.} Currently, transformers have achieved state-of-the-art performance on a wide range of downstream NLP tasks. Transformers are a type of deep contextualized language model that typically operate using stacked attention layers. These models are capable of capturing long-range dependencies in language by attending to relevant words in the input sequence at each layer, allowing them to achieve impressive performance on a variety of NLP tasks \citep{vaswani2017attention} (briefly explained in Section~\ref{sec:gap}). Many of these models, such as BERT \citep{devlin2018bert}, undergo a two-phase process, consisting of pretraining and fine-tuning. During pretraining, the model is trained on a masked language modeling task, whereby certain tokens within the input sequence are masked, and the model is trained to predict the masked tokens. This process enables the model to acquire a deep understanding of the underlying linguistic structure of the language, including its syntax and semantics. In the subsequent fine-tuning phase, the pretrained model is further optimized for performance on downstream tasks, such as sentiment classification \citep{socher2013recursive} and paraphrase detection \citep{dolan2005automatically}. By fine-tuning the model on these specific tasks, it can be tailored to achieve state-of-the-art results, leveraging the powerful contextualization capabilities of the Transformer architecture. For instance, in the case of sentiment classification, a new multi-layer perceptron (MLP) could be appended to the encoded output of the main model to generate a binary decision for a given sentence. The parameters of both the added MLP and the pretrained model can then be fine-tuned using the available training data for sentiment classification. With the abundance of training data, the vast amount of textual context, and the powerful capabilities of the Transformer architecture, one could argue that visual grounding does not offer any additional information for solving current NLP tasks \citep{tan-bansal-2020-vokenization}. 

Despite the arguments against the necessity of visual grounding for transformer-based language models, we are curious about the potential benefits of our simple grounding approach. To explore this possibility, we incorporated our approach with BERT \citep{devlin2018bert}, one of the pioneering transformer models for sentence-level natural language understanding tasks. BERT has been pre-trained on a vast corpus of English text, including English Wikipedia\footnote{\url{https://en.wikipedia.org/wiki/English\_Wikipedia}} and BookCorpus \citep{Zhu_2015_ICCV}, a collection of 11,038 unpublished books. We carry out new experiments to compare the performance of visually grounded BERT and purely textual BERT on sentence-level NLP tasks. To clarify, in our baseline model, fixed FastText or GloVe vectors serve as the input to the $\textbf{M}$ mapping. However, in our new model, these vectors are replaced by vectors generated through BERT encoding. The BERT encoder marks the beginning and end of the input with `[cls]' and `[sep]' tokens (as shown in Figure~\ref{fig:bert}) and outputs a fixed-dimensional vector for each token. Therefore, we can treat it as a word-embedding model. Given a sentence ($S_j=[w_1,w_2, \cdots, w_n]$) with $n$ words, the BERT encoder outputs ($T_j=[t_1,t_2, \cdots, t_n]$), where $t_i$ represents the contextualized encoding of the word $w_i$.


When used for classification tasks, the BERT engine is coupled with a multi-layer-perceptron network generating the final output. As shown in Figure~\ref{fig:bert}, similar to our proposed model, we train a linear mapping $M$ followed by an LSTM encoder to predict an image vector given its caption. After the training phase (see the lower box), for each classification task, the pre-trained model has to be fine-tuned. For this step, an MLP is added on top of the mapping $M$ for fine-tuning on the downstream task (see the upper box). In the fine-tuning phase, the `[cls]' tokens encode the given input through multiple attention layers and the rest of the tokens are discarded \citep{devlin2018bert}. In a nutshell, our approach adds the learned alignment $M$ between the pre-trained BERT encoder and its classifier. This alignment is applied to the BERT encoding to align its final representation to vision without deteriorating its textual information.\\

\noindent \textbf{Evaluation:} We fine-tuned and evaluated our pre-trained grounded BERT on the General Language Understanding Evaluation (GLUE) benchmark\footnote{https://gluebenchmark.com/} \citep{wang2018glue} implemented in the Huggingface\footnote{https://huggingface.co/} library \citep{wolf2019huggingface}. GLUE is widely regarded as a comprehensive  evaluation suite for natural language understanding models that reflect a wide range of the complexity and diversity of human language comprehension. It consists of nine natural language understanding tasks: single-sentence tasks, SST-2 \citep{socher2013recursive} and CoLA \citep{warstadt2019neural}; paraphrasing and similarity tasks, MRPC \citep{dolan2005automatically}, QQP\footnote{https://quoradata.quora.com/First-Quora-Dataset-Release-Question-Pairs}, and STS-B \citep{cer2017semeval}; natural language inference tasks, RTE \citep{wang2018glue}, QNLI \citep{rajpurkar2016squad}, MNLI \citep{williams2017broad}, and WNLI \citep{levesque2012winograd}. In what follows, we briefly explain the GLUE tasks used in our experiments.\\

\noindent \textbf{SST-2:}The Stanford Sentiment Treebank compiles a set of sentiment annotations from movie reviews. It includes a total of 215,154 phrases each annotated by 3 human annotators. Each sample is assigned to one of the following five labels: neutral, slightly neutral, moderately positive, or positive. SST-5 or SST fine-grained refers to the corpus with all 5 labels. SST-2 however consists of binary labels only. The Negative class indicates negative or slightly negative and the positive class indicates somewhat positive or positive. The neutral sentences are discarded in SST-2 resulting in 70,042 overall samples. Examples of positive and negative sentences are \textit{`that loves its characters and communicates something rather beautiful about human nature'} and \textit{`that 's far too tragic to merit such superficial treatment'} accordingly.\\\\
\noindent \textbf{CoLA:} The Corpus of Linguistic Acceptability is an English acceptability evaluation dataset. It consists of 10,657 sentences from 23 linguistics publications, expertly annotated for acceptability (grammaticality) by their original authors into positive and negative classes. Some negative examples are: \textit{`The professor talked us'}, \textit{`They made him to exhaustion'}, and \textit{`The witch went into the forest by vanishing'}.\\\\
\noindent \textbf{MRPC:} The Microsoft Research Paraphrase Corpus is a set of sentence pairs retrieved from online news sources. MRPC includes 5801 sentence pairs, each labeled by human judges as to whether the pair constitutes a paraphrase. This task is also known as paraphrase detection. Examples from this dataset are, \textbf{positive:} (\textit{`About 130,000 U.S. troops remain in Iraq , with others deployed in Afghanistan , South Korea and elsewhere.'}, \textit{`About 130,000 US soldiers remain in Iraq , with others serving in Afghanistan, South Korea , Japan , Germany and elsewhere.'}); \textbf{negative:} (\textit{`The Embraer jets are scheduled to be delivered by September 2006.'},  \textit{`The Bombardier and Embraer aircraft will be delivered to U.S. Airways by September 2006.'}).\\\\
\noindent \textbf{QQP:} The Quora Question Pairs, is a collection of question pairs from the question-answering website Quora. The task is identical to that of MRPC. the QQP, however, is much larger, it compiles a set of 400k question pairs each with a binary label indicating the semantic equivalence of the question pair.\\\\
\noindent \textbf{STS-B:} The Semantic Textual Similarity Benchmark is a set of sentence pairs compiled from captions for videos and images, natural language inference data, and news headlines. It consists of 8628 sentence pairs with each pair annotated by humans with a similarity score ranging from 1 to 5. The task is to predict the similarity score of a given pair as a real-valued number. For example, (\textit{`A woman is dancing.', `A man is talking'}) has a score of $0$ and (\textit{`A small dog is chasing a yoga ball', `A dog is chasing a ball'}) has a score of $4$.\\\\
\noindent \textbf{RTE:} Recognizing Textual Entailment is the task of modeling a directional relation between two sentences. The relation holds whenever the truth of the second sentence is entailed by the first one. For instance, \textit{`a dog is jumping for a Frisbee in the snow'} entails \textit{`An animal is outside in the cold weather, playing with a plastic toy.'} but contradicts \textit{`a cat washed his face and whiskers with his front paw.'}. The RTE dataset consists of 5767 pairs, extracted from news and Wikipedia text, each with a binary label.\\\\
\noindent \textbf{QNLI:} The Stanford Question Answering Dataset consists of question-paragraph pairs. One of the sentences in the paragraph (drawn from Wikipedia) contains the answer to the question in the given sample. Questions are written by human annotators. To convert this task into a sentence pair classification one, \citet{wang2018glue} constructed a pair between each question and each
sentence in the corresponding paragraph, and discarded pairs with low lexical overlap between the
question and the context (paragraph) sentence. The task is to predict whether the context sentence contains the answer to the question. This dataset contains 115,699 question-sentence pairs each annotated with a binary label. Examples from this dataset are, \textbf{positive:} (\textit{`When is the term 'German dialects' used in regard to the German language?', 
`When talking about the German language, the term German dialects is only used for the traditional regional varieties.'}), \textbf{negative: }(\textit{`In what century was the church established at the location?', `Construction of the present church began in 1245, on the orders of King Henry III.'})\\\\
\noindent \textbf{MNLI:} The Multi-Genre Natural Language Inference is a dataset of 431,992 sentence pairs with entailment annotations. Given a pair of premise-hypothesis sentences, the task is to predict whether the premise entails the hypothesis (entailment), contradicts the hypothesis (contradiction), or neither (neutral). The premise sentences are
gathered from different sources including government reports, transcribed speech, and fiction. There are two versions of the validation set, matched  and mismatched. The former contains samples in the same domain as in the training set while the latter contains cross-domain samples. We evaluate our model on both sets.\\\\

\noindent \textbf{Implementation Details:} We used the \textit{bert-base-cased} version of BERT \citep{devlin2018bert} in our experiments. `\textit{base}' refers to the size of the model in terms of the number of training parameters. There are three versions of BERT: \textit{small, base,} and \textit{large}; `cased' indicates that the model distinguishes between upper-cased and lower-cased letters.  For training, we used the Microsoft COCO 2017 dataset \citep{lin2014microsoft}. The alignment $M$ maps a BERT token $t_i\in \mathbb{R}^{768}$ to $g_i\in \mathbb{R}^{1024}$. Each LSTM layer contains $1024$ units. A single-layer neural network with a linear activation function (a linear layer) is applied on top of the LSTM to predict the image vector $I_j\in \mathbb{R}^{2048}$. We trained the model on image-caption pairs for 10 epochs using the AdamW optimizer~\citep{loshchilov2017decoupled} with the learning rate set to  $5e^{-5}$ and a batch size of $64$. For fine-tuning on the GLUE benchmark, we followed the huggingface guidelines\footnote{https://github.com/huggingface/transformers/tree/master/examples/pytorch/text-classification} and fine-tuned the model on each downstream task for 5 epochs with a batch size of 32 and a learning rate of $2e^{-5}$.\\

\begin{table*}[ht]
\centering
\begin{adjustbox}{width=1\textwidth}
\begin{tabular}{lccccccccc}
\hline \textbf{Model/Data} & \textbf{CoLA} & \textbf{MRPC} & \textbf{QNLI} & \textbf{QQP} & \textbf{RTE} & \textbf{SST-2} & \textbf{MNLI} & \textbf{STS-B} & \textbf{Mean Score}\\
Train Size (K) & 8.5  &3.6  &104 &364 &2.5 &67 &392 & 5.7 & - \\ \hline

Textual-BERT    & 59.05 & 84.31/89.15  & 91.08 & 90.76/87.53 & 67.15 &91.2 & 83.34/83.83 & 87.13/87.00  & 81.74 \\
1-LFM-GBERT  &60.07 & 84.31/89.11 & 91.00 & 90.82/87.63 & 63.54 & \bftab 92.43 & 83.86/83.52 &88.83/88.49 &81.86 \\  
2-LFM-GBERT     & 61.58 & 85.29/89.58 & 91.47 & 90.71/87.44 & 67.15 & 92.09 & 83.78/83.66 &88.44/88.04 &82.56 \\
3-LFM-GBERT      & 60.62 & 84.56/89.44 & 90.92 & 90.70/87.46 & \bftab 68.23 & 92.32 & 83.84/83.48 & 88.02/87.67 &82.40  \\
\hline
2-LTM-GBERT &   \bftab 61.62 & \bftab 86.27/90.51 &91.12  & 90.73/87.46 & 67.15 & 92.20 & 83.73/83.71 & \bftab 89.12/88.74 & \bftab 82.74 \\
\hline

\end{tabular}
\end{adjustbox}
\caption{\label{tab:bert} Validation scores on the GLUE benchmark using textual BERT and visually grounded BERT (\textit{*\_GBERT}). Visual grounding seems to improve the generalization of the model when training data is limited (e.g., MRPC and CoLA). However, large volumes of training data compensate for visual grounding (see the scores of QQP and MNLI). \textit{accuracy/F1\_scores} are reported for QQP and MRPC, \textit{Pearson/Spearman} correlations are reported for STS-B, and accuracies for \textit{matched/mismatched} sets are reported for MNLI. For the other tasks, accuracy is reported. Numbers in bold indicate obvious improvements over textual BERT.}

\end{table*}

\begin{table*}[ht]
\centering
\begin{adjustbox}{width=1\textwidth}
\begin{tabular}{lccccccccc}
\hline \textbf{Model/Data} & \textbf{CoLA} & \textbf{MRPC} & \textbf{QNLI} & \textbf{QQP} & \textbf{RTE} & \textbf{SST-2} & \textbf{MNLI} & \textbf{STS-B} & \textbf{Mean Score}\\
\hline

Textual-BERT     & 73.92 & 68.38/81.22    & 52.48  & 63.18/00.00  & \bftab 52.35 & 82.34 & 36.40     & 22.70/09.78& 56.47 \\
Grounded-BERT     & \bftab 77.85 & 68.38/81.22 & \bftab 54.42 & \bftab 67.21/48.63  &48.38   &85.55  & \bftab 42.25  & \bftab 47.80/47.29  & 61.48  \\
\hline

\end{tabular}
\end{adjustbox}
\caption{\label{tab:bert_prob} Validation scores on the GLUE benchmark by employing a linear probe on textual BERT and visually grounded BERT. The visually grounded vector space provides richer semantic representations, leading to improved language understanding on a majority of the tasks. Numbers in bold indicate significant differences in performance (p values $<0.05$).}

\end{table*}

\noindent \textbf{Results:} Table~\ref{tab:bert} reports the validation scores across the GLUE datasets. The WNLI dataset was excluded from the list following \citet{devlin2018bert} due to inconsistent results. We carried out our grounding experiments with different numbers of LSTM layers. In Table~\ref{tab:bert}, \textit{n-LFM-GBERT} indicates the grounded BERT with \textit{n} layers of LSTMs and frozen (weights are kept unchanged during training) mapping $\mathbf{M}$ while fine-tuning on downstream tasks. The idea behind freezing the mapping (alignment) $\mathbf{M}$ while fine-tuning the BERT encoder and the classifier on a particular task is to guide (force) the output representations of BERT to follow the visual alignment. This might then guide the model to a better feature space for solving the task. Considering the mean score, the grounded model with 2-layer-LSTMs (\textit{2-LFM-GBERT}) outperforms the textual BERT by almost $1 \%$, highlighting the potential benefits of visual grounding. Moreover, we also fine-tuned the alignment $\mathbf{M}$ of the best model (\textit{2-LFM-GBERT}) for each particular task along with BERT encoder and the classifier, denoted as \textit{2-LTM-GBERT}, this model further improves the results. Although the improvements achieved through visual grounding in our experiments are marginal compared to those obtained through grounded word embeddings, the results presented in the table provide valuable insights. Notably, for datasets with limited training data, such as CoLA and MRPC, visual grounding appears to provide an advantage, as indicated by the bold numbers in the table. However, for larger datasets such as QQP and MNLI, the results are almost identical for both grounded and textual BERT models. These findings suggest that visual grounding improves the generalization of transformers when training data is limited. Nonetheless, they also demonstrate that a substantial amount of textual training data, combined with meticulous fine-tuning of models, can compensate for the relatively simple visual grounding approaches used in our study when tested on the GLUE benchmark. In accordance with our prior word embeddings experiments, we conducted a t-test comparing the results of \textit{textual BERT} to those of \textit{Grounded BERT}, more specifically \textit{2-LTM-GBERT}. The statistical test indicated that the observed enhancements in performance were \textbf{not} statistically significant. Nevertheless, when compared to the process of human language acquisition, these textual language models exhibit significant inefficiencies, requiring exposure to vast amounts of training data and computational resources to achieve satisfactory results \citep{strubell-etal-2019-energy}. The BERT model for instance, despite being pre-trained on an extensive corpus of over 3 billion tokens, still requires meticulous fine-tuning for each individual task, which raises doubts about the efficacy of large language models and the potential usefulness of visual grounding in this regard.\\
\hspace*{\parindent}
In light of these concerns, we conducted an investigation to determine whether fine-tuning the model would obscure 
improvements in the overall quality of embeddings due to 
visual grounding. In other words, fine-tuning the models might diminish the differences between them, as the learned parameters are tailored to the specific downstream task, potentially obscuring the benefits of visual grounding. For this aim, we designed a new experiment whereby we skipped the fine-tuning phase and conducted a comparative analysis of the semantic spaces of Textual BERT and Grounded BERT models. Despite the adverse impact of skipping fine-tuning on the results, this experimental approach enables us to juxtapose the semantic space of the two models more accurately and identify potential subtle differences between them, with a particular focus on the influence of visual grounding. To compare the semantic space of Grounded BERT and Textual BERT for each specific task within the GLUE benchmark, we employ a technique called \textit{linear probing}. In this technique, only a linear classifier such as logistic regression is trained on top of pre-trained representations of a model, in order to measure the quality of the learned representations for particular downstream tasks \citep{reif2019visualizing}. For tasks involving pairs of sentences, a linear probe is trained with the cosine similarity between the representations of the two sentences. For instance, consider the task of paraphrase detection using the MRPC dataset, which involves predicting whether a given pair of sentences are semantically equivalent. In our probing setup, the two sentences, $s_1$ and $s_2$,  are first encoded separately by Grounded BERT and Textual BERT, resulting in two vectors, $v_1$ and $v_2$, representing each sentence. We then determine the semantic similarity of the two sentences by calculating the cosine similarity between the two vectors:
$ score (v_1,v_2)= 1 -\frac{v_1.v2}{\lVert v_1 \rVert \lVert v_2 \rVert} $.
After encoding the sentences and calculating cosine similarities between the two vectors, a logistic regression model (the probe) is trained using the cosine similarities as inputs and binary classification labels as outputs. Following training, the trained linear probe is applied to predict the labels of the validation set. The rest of the evaluation procedure is identical to the previous section. If one of the models' representations is better suited for this task, we expect to observe higher performance,  indicating better classification boundaries and more refined clusters in the semantic space of the model. \\
\hspace*{\parindent}
The evaluation results of probing are reported in Table~\ref{tab:bert_prob}. Grounded BERT demonstrates significant improvements over textual BERT leading to the enhancement of the mean score by $5\%$. This shows that visual grounding enriches language representations across a wide range of abstract language understanding tasks. Surprisingly, the accuracy on \textit{CoLA} dataset, is higher than when the whole model is fine-tuned (see Tabel~\ref{tab:bert}). This might be due to the nature of the task. Since negative samples contain ungrammatical sentences, they might inherently be well separated from correctly grammatical sentences in the vector space. Hence, fine-tuning the parameters of BERT with a small set of ungrammatical sentences might be detrimental to model performance. This further confirms the inefficiencies of large language models and their need to devour a huge amount of annotated data to achieve desirable performance. We further performed a t-test between the prediction of the two models, exhibiting statistically significant differences between the performance of the two models on the majority of the tasks. Bold numbers in Table~\ref{tab:bert_prob} indicate p values $< 0.05$.

Overall, these insights highlight the potential of visual grounding even for highly advanced NLP techniques. Our findings suggest that visual grounding has the potential to learn task-agnostic language representations, leading to reduced computational costs and textual resources. This paves the way for future research on building cognitively plausible language learning frameworks where the learning process leverages different modalities such as visual cues and gestures \citep{Smith:Gasser:2005, iverson2005gesture}, making the learning both effective and cognitively plausible.

\section{Grounding for smaller datasets}
\label{sec:grounding_small_ds}
{Thus far, our grounding approach has been shown to be effective in conjunction with pre-trained word embedding models and advanced sentence-level language models, when training data for a given downstream task is scarce. In both cases, however, large amounts of textual training data from different domains have been utilized.  The amount of training data plays a big role in shaping performance on downstream tasks \citep{beltagy2019scibert, lee2020biobert}, and in general is an important determinant of the quality of industrial word embeddings \citep{wang2019evaluating, elekes2018resources, johns2022content}. This section details two concluding experiments that address the question of whether visual grounding is also beneficial for embeddings calculated from much more modest training data.  As human lexical acquisition develops rapidly on the basis of restricted amounts of training data, a solid improvement due to visual grounding even under limited exposure would provide support for the possibility that human learning also benefits from visual grounding.} 

We, therefore, trained the GloVe model from scratch on two small and different training corpora and measured the improvements of our grounding approach on each corpus using the word similarity benchmarks (see section ~\ref{sec:evaluation}). Initially, we obtained textual embeddings by training on two distinct corpora: TASA and Text8. TASA \citep{zeno1995educator} has served as a training corpus for, e.g., Latent Semantic Analysis \citep{Landauer1999LatentAlone}. Text8 is a small corpus sampled from Wikipedia to allow quick testing of language models.\footnote {\url{https://cs.fit.edu/~mmahoney/compression/textdata.html}} Our best grounding model (see Section~\ref{sec:proposed_method}) is then applied to the textual embeddings to obtain visually grounded embeddings.  Table~\ref{nlp_limited_data} reports the comparison between textual embeddings and grounded embeddings for both corpora. Our grounded approach (TASA-G and Text8-G) consistently improves on top of textual embeddings (TASA-T and Text8-T) despite the small size of these corpora and the very different nature of the training corpora.{We further confirmed the statistical significance ($p \leq 0.0008$) of the performance improvements observed by conducting t-tests on both datasets.} The robustness of our grounding method for word-based embeddings holds not only across a wide range of tasks, but also for different amounts of training data, providing a firm basis for expecting grounded embeddings to provide improved precision to studies of human cognition that make use of embeddings.

\begin{table*}[ht]
\centering
\begin{adjustbox}{width=1\textwidth}
\begin{tabular}{lccccccccc}
\hline \textbf{Model} & \textbf{RW} & \textbf{MEN} & \textbf{WSim} & \textbf{MTurk} & \textbf{SimVerb} & \textbf{SimLex} & \textbf{Mean}\\
&  &  & \textbf{353} & \textbf{771} & \textbf{3500} & \textbf{999} & \\ \hline

TASA-T   & 2.4 & 37.2 & 33.1 & 35 & 8.5 & 10.8 & 21.7 \\
TASA-G &   \bftab 6.7 & \bftab 42.5 & \bftab 37.1 & \bftab 37.5 & \bftab 13.1 & \bftab 17.1 &  \bftab 25.7 \\
\hline
Text8-T     & 8.1 & 47.9 & 45.9 &  45 & 8.3 & 16.6 & 28.63\\
Text8-G   & \bftab 13.5 & \bftab 51.2 & \bftab 51.8 & \bftab 49.1 & \bftab 10.5 & \bftab 20.7 & \bftab 32.8\\
\hline
\end{tabular}
\end{adjustbox}
\caption{\label{nlp_limited_data} Comparison of our grounded embeddings (*-G) to textual embeddings (*-T) on limited training data. GloVe algorithm was trained on TASA and Text8 corpus separately from scratch. Significant improvements are achieved by visual grounding despite limited training data. Numbers in bold indicate significant differences in performance (p values $\leq 0.0008$, t-tests).}

\end{table*}



\section{Discussion and Conclusion}


In this study, we designed a visual grounding framework that effectively produces visually grounded word representations for all types of words from different kinds of embeddings. Our approach, apart from its simplicity, shows excellent generalization, as evidenced by its success on a variety of human-annotated similarity and relatedness tasks, including those involving unseen abstract and concrete words. We have made both the grounded embeddings and our framework publicly available. We further designed a series of experiments to shed light on the following research questions. \\

{
\noindent \textbf{Visual grounding for abstract words:} Our approach employs a visual grounding pathway that is acquired during the process of grounding concrete words, which enables the indirect grounding of abstract concepts. Our study's results lend support to the indirect grounding theory, which posits that concrete words are directly grounded while abstract words are indirectly grounded through language \citep{howell2005model, louwerse2011symbol,hoffman2018concepts}. {Despite being trained on image captions within which concrete nouns far outnumber abstract nouns, our approach produces more refined clusters of both concrete and abstract words}, highlighting the framework's ability to capture the subtle nuances in the semantics of different word types across a wide range of human-annotated word collections.
}

\textbf{Bridging language to vision:} We investigated various strategies of bridging language (here crudely represented as word/sentence embeddings) with vision. Our experiments support the following conclusions. 


First, textual word embeddings benefit from vision the most when they are aligned with vision as opposed to being merged. Our alignment strategy enables the textual embeddings to incorporate real-world knowledge through images without compromising the statistical knowledge gained from textual corpora. We showed by example that allowing too much visual information will overwhelm the textual embeddings. Injecting too much visual knowledge into the embeddings benefits concrete words while diminishing the performance on modeling abstract words. {This trade-off may be due to the distinct cognitive processing of abstract and concrete words, which engage overlapping but separate brain regions \citep[see][for reviews]{Montefinese2019, Mkrtychian2019}. Therefore, the right balance between concreteness and abstractedness represented in our experiments by visual properties of images and statistics of textual corpora is vital.}

Our second key finding is that textual context plays an important role in grounding isolated word embeddings. Our results demonstrate that linking word embeddings with vision in the absence of textual context leads to a significant distortion of the semantic space. We believe one reason is that word vectors still need to be aware of the textual context they occur in when they are being coupled with their corresponding visual information in images. Moreover, given that images are a highly complex and rich source of information, a single word cannot capture their full semantic richness. Our grounding framework, therefore, aligns word vectors with their corresponding images while simultaneously preserving information about their textual context, thereby enhancing the overall efficacy of the grounding process.


{\noindent \textbf{Benefits and upper bound of visual grounding:} Our study has demonstrated that visual grounding is highly advantageous for both concrete and abstract words. However, our analyses have also revealed that visual grounding is particularly beneficial in cases where textual embeddings struggle, such as when modeling highly abstract verbs or rare words. Conversely, in benchmarks consisting mostly of concrete words, the improvement from grounding is less pronounced. These findings dovetail well with the observation that the meanings of concrete words are more stable and reliable compared to those of abstract words across different textual word embeddings \citep{pierrejean2019investigating}.\\
\hspace*{\parindent} It has been shown that infants' ability for processing abstract words emerges later after they have established a solid grounding in concrete concepts \citep{bergelson2013acquisition,bergelson20126}. Furthermore, many abstract concepts build on metaphors that themselves are rooted in concrete experiences \citep{Lakoff:Johnson:1980,Langacker:1987}.
This finding suggests a possible high-level explanation of why abstract words benefit from visual grounding of concrete words: Abstract words are scaffolded on the foundations of concrete words.  Visual grounding contributes to a more precise approximation of these foundations, and this in turn enables a recalibration of the superstructure of abstract words. Our findings thus pave the way for future research on whether visual grounding alleviates the instability problem of abstract concepts \citep{pierrejean2019investigating}.} \\
\hspace*{\parindent} \textbf{Visual grounding and corpus size:} 
The embeddings used in current NLP are derived from corpora comprising billions of words.  An examination of the extent to which visual grounding helps improve state-of-the-art sentence-level NLP models built on such huge resources revealed only modest improvements. Specifically, a comparison of  a visually grounded version of the well-known BERT model \citep{devlin2018bert}  with a standard textual version of BERT on common evaluation benchmarks showed that visual grounding yields considerable improvements only when training data is limited. However, when using large volumes of textual data and meticulous parameter-tuning, the performance of the visually grounded and textual models becomes almost identical. Apparently, huge volumes of textual context in combination with subsequent powerful fine-tuning algorithms compensate for visual grounding, at least on current downstream NLP tasks. 

Although visual grounding is not necessary for language models that have access to volumes of data that far surpass what individual speakers can ever encounter, we have shown that when embeddings are trained on small corpora, visual grounding leads to substantial improvements. 

Since we as humans are never exposed to the amount of textual data digested by current language models, but still master our first language at a very early age, enriching current models for lexical semantics with vision is a first step forward in the direction of developing cognitively more plausible representations for word meaning.

\section*{Acknowledgements}
Funded by the Deutsche Forschungsgemeinschaft (DFG, German Research Foundation) under Germany’s Excellence Strategy – EXC number 2064/1 – Project number 390727645, as well as by the German Federal Ministry of Education and Research (BMBF): Tübingen AI Center, FKZ: 01IS18039A. The authors thank the International Max Planck Research School for Intelligent Systems (IMPRS-IS) for supporting Hassan Shahmohammadi.

\section*{Open Practices Statement}
Code for acquiring grounded word embeddings and two sets of ready-to-use grounded embeddings are available at \url{https://github.com/Hazel1994/Visually_Grounded_Word_Embeddings_2}.
\bibliography{sn-bibliography}


\end{document}